\documentclass{bmvc2k}

\usepackage{booktabs}
\usepackage{amsmath,amsfonts}
\usepackage{algorithm}
\usepackage{algpseudocode}
\usepackage{multirow}
\usepackage{bm}

\usepackage{enumitem}
\usepackage[export]{adjustbox}
\usepackage{subfigure}
\usepackage{graphicx}
\usepackage{ulem}
\usepackage{xspace}
\usepackage{soul}
\usepackage{wrapfig}
\usepackage{sidecap}

\usepackage{microtype}

\usepackage{todonotes}

\newcommand{\ourmethod}{SoftClu\xspace}
\title{Data Augmentation-free Unsupervised Learning for 3D Point Cloud Understanding}

\addauthor{\footnotesize Guofeng Mei}{guofeng.mei@student.uts.edu.au}{1}
\addauthor{\footnotesize Cristiano Saltori}{cristiano.saltori@unitn.it}{2}
\addauthor{\footnotesize Fabio Poiesi}{poiesi@fbk.eu}{3}
\addauthor{\footnotesize Jian Zhang}{jian.zhang@uts.edu.au}{1}
\addauthor{\footnotesize Elisa Ricci}{e.ricci@unitn.it}{2}
\addauthor{\footnotesize Nicu Sebe}{niculae.sebe@unitn.it}{2}
\addauthor{\footnotesize Qiang Wu}{qiang.wu@uts.edu.au}{1}

\addinstitution{
Global Big Data Technologies Centre\\
 University of Technology Sydney\\
 Sydney, Australia
}
\addinstitution{
Multimedia and Human Understanding Group\\
 University of Trento\\
 Trento, Italy
}
\addinstitution{
Technologies of Vision Lab\\
 Fondazione Bruno Kessler\\
 Trento, Italy
}

\runninghead{Student, Prof, Collaborator}{BMVC Author Guidelines}

\begin{document}\sloppy
%-------------------------------------------------------------------------
\maketitle
\vspace{-0.8cm}
\begin{abstract}
Unsupervised learning on 3D point clouds has undergone a rapid evolution, especially thanks to data augmentation-based contrastive methods. However, data augmentation is not ideal as it requires a careful selection of the type of augmentations to perform, which in turn can affect the geometric and semantic information learned by the network during self-training. 
To overcome this issue, we propose an augmentation-free unsupervised approach for point clouds to learn transferable point-level features via soft clustering, named \ourmethod. 
\ourmethod assumes that the points belonging to a cluster should be close to each other in both geometric and feature spaces. This differs from typical contrastive learning, which builds similar representations for a whole point cloud and its augmented versions. 
We exploit the affiliation of points to their clusters as a proxy to enable self-training through a pseudo-label prediction task. 
Under the constraint that these pseudo-labels induce the equipartition of the point cloud, we cast \ourmethod as an optimal transport problem.
% which can be solved by using an efficient variant of the Sinkhorn-Knopp algorithm. 
We formulate an unsupervised loss to minimize the standard cross-entropy between pseudo-labels and predicted labels. 
Experiments on downstream applications, such as 3D object classification, part segmentation, and semantic segmentation, show the effectiveness of our framework in outperforming state-of-the-art techniques~\href{https://github.com/gfmei/softclu}{[code]}.
\end{abstract}

%-------------------------------------------------------------------------
%%%%%%%%%%%%%%%%%%%%%%%%%%%%%%
%%%%%%%%%%%%%%%%%%%%%%%%%%%%%%%%%%%%%%%%%%%%%%%%%%%%%%
%%%%%%%%%%%%%%%%%%%%%%%%%%%%%%%%%%%%%%%%%%%%%%%%%%%%%%
%%%%%%%%%%%%%%%%%%%%%%%%%%%%%%%%%%%%%%%%%%%%%%%%%%%%%%
\vspace{-0.8cm}
\section{Introduction}
\vspace{-0.2cm}
The rapid progress of 3D capturing devices, such as 3D laser scanners and depth sensors, led to convenient and effective ways to process 3D data, which, if combined with RGB images, can further improve the understanding of environments.
Applications like robotic navigation \cite{biswas2012depth,Zhou2022}, autonomous driving \cite{li2020deep} and exploration \cite{wang2020}, and augmented and virtual reality \cite{park2008multiple} are among the major reasons for a higher attention toward 3D data understanding.

Learning discriminative and transferable point cloud features is a crucial problem in the area of 3D shape understanding \cite{lin2021object,poiesi2022}, as it allows efficient training of downstream tasks, such as 
object detection \cite{shi2020pv} and tracking~\cite{yin2021center}, 
segmentation \cite{xu2020geometry}, 
reconstruction~\cite{yuan2018pcn}, 
classification \cite{qi2017pointnet}, 
and registration \cite{huang2020feature,mei2021point,mei2022overlap}.
Therefore, learning from unlabeled or partially labeled data to alleviate human labeling efforts is an emerging research topic in point cloud understanding.
Along this line, unsupervised representation learning is an attractive, yet potent, alternative approach to learning features without human intervention \cite{eckart2021self}. 

% Unsupervised representation learning aims at obtaining transferable features through abundant unlabeled data \cite{xie2021unsupervised}.
Unsupervised learning approaches can be broadly categorized as generative or discriminative \cite{grill2020bootstrap}. 
The former includes self-reconstruction or auto-encoding \cite{yang2018foldingnet}, generative adversarial network \cite{sarmad2019rl}, and auto-regressive \cite{sun2020pointgrow} methods.
These methods can map an input point cloud into a global latent representation \cite{rao2020global,shi2020unsupervised}, or a latent distribution in the variational case \cite{hassani2019unsupervised,han2019multi} through an encoder and then attempt to reconstruct the input by a decoder. 
Generative methods can be effective to model high-level and structural properties of the input point clouds. However, because they are sensitive to Euclidean transformations, they typically assume that all 3D objects have the same pose in a given category \cite{sanghi2020info3d}.

Unlike generative methods, discriminative methods learn to predict or discriminate augmented versions of the input. 
These methods can yield rich latent representations for downstream tasks \cite{wang2020unsupervised}. 
Examples of these include contrastive methods \cite{du2021self,rao2020global,sanghi2020info3d}, which have shown remarkable results for unsupervised representation learning. 
Contrastive methods also promote learning of rotation-invariant representations via data augmentation \cite{poursaeed2020self}. 
Typically, these algorithms require several negative samples and heavily depend on the selection criteria to mine negatives \cite{grill2020bootstrap,chen2021exploring}. 
Often, they require large batch sizes, memory banks, or customized strategies to retrieve informative pairs \cite{grill2020bootstrap}. 
Moreover, it is somewhat unclear what constitutes an effective semantics-preserving data augmentation strategy given that a point cloud is typically defined as a set of 3D coordinates. 
Any disturbance of the original geometry, such as cropping or view-based occlusions, could potentially degrade its semantics \cite{eckart2021self}; for example, random crops of a point cloud may correspond to different objects and introduce inconsistent learning signals.
For this reason, contrastive approaches need humans to carefully design combinations of data augmentations to learn informative representations.
% Local geometric features, e.g., those encoded by a point through the contribution of its neighbors, have shown to be discriminative for 3D tasks \cite{qi2017pointnet, qi2017pointnet++}. 
On the other hand, training on whole object instances can lead to the learning of global representations, which in turn can produce fewer discriminant representations as local geometric differences may be disregarded \cite{rao2020global, han2019multi, xie2020pointcontrast}.
Therefore, our first motivation is to design a data augmentation-free unsupervised learning approach to avoid the inconvenience of building chains of ad-hoc combinations of data augmentations. 
Second, we develop an unsupervised method that is not based on global features but instead can optimize local features, which facilitates the network to learn 3D spatial geometric information of point clouds.

In this paper, we propose an unsupervised method to learn informative point-level representations of 3D point clouds without using data augmentation. 
Our framework learns cluster affiliation scores to softly group the 3D points of each point cloud into a given number of geometric partitions, i.e.~through soft clustering. We learn point-level feature representations by minimizing the standard cross-entropy of a single equation, which is the result of an EM-like algorithm \cite{moon1996expectation}. 
The Expectation step employs an optimal transport \cite{peyre2019computational} based clustering algorithm to generate point-level pseudo-labels, i.e.~focusing on local geometric information.
In particular, we softly label points based on their distance from the centroids in both feature and geometric spaces, with the constraint that labels partition data in equally-sized subsets (uniform distribution).
Optimal transport serves as a potent means for comparing probability distributions with each other, as well as for producing optimal mappings to minimize distances \cite{peyre2019computational}.
The Maximization step adapts the E-step for a point-to-cluster loss to optimize the metric learning network.
Our approach learns the partitioning network itself and softly assigns points into geometrically coherent overlapping clusters, {overcoming the weakness of conventional GMM and K-means that involve expensive iterative procedures.} In doing so, we avoid data augmentation that might degrade its geometric coherency and thus its semantic information. 
Our approach is inspired by DeepCluster~\cite{caron2018deep}, SeLa~\cite{asano2020self} and 
SwAV~\cite{caron2020unsupervised} but it differs from them, as they implement clustering in the feature space at the instance level, they use data-augmentation, and they may degrade the geometric information when used with 3D data.
% \fabiocomment{"Working only with a well-balanced set of single-class object-centric data." -> this comment is not clear to me.}
We show that pre-training on datasets using \ourmethod can improve the performance of a range of downstream tasks and outperforms the recent state-of-the-art methods without any data augmentation and also across different domains.

\noindent To summarize, our contributions are:
\begin{itemize} %[noitemsep,topsep=0pt,leftmargin=*]
	\item We propose a data augmentation-free unsupervised method, which does not rely on data augmentations, negative pair sampling, and large batches, to learn transferable point-level features on a 3D point cloud;
	\item We extend the pseudo-label prediction to an optimal transport problem, which can be efficiently solved by using an efficient variant of the Sinkhorn-Knopp~\cite{cuturi2013sinkhorn} algorithm;
	\item We conduct thorough experiments, and \ourmethod achieves state-of-the-art performance without having to use data augmentation.
\end{itemize}
%%%%%%%%%%%%%%%%%%%%%%%%%%%%%%

%%%%%%%%%%%%%%%%%%%%%%%%%%%%%%
%%%%%%%%%%%%%%%%%%%%%%%%%%%%%%%%%%%%%%%%%%%%%%%%%%%%
%%%%%%%%%%%%%%%%%%%%%%%%%%%%%%%%%%%%%%%%%%%%%%%%%%%%
%%%%%%%%%%%%%%%%%%%%%%%%%%%%%%%%%%%%%%%%%%%%%%%%%%%%
\section{Related Work}
In this section, we briefly review existing works related to unsupervised learning on point clouds, which can be classified into two categories: generative and discriminative methods. 
%%%%%%%%%%%%%%%%%%%%%%%%%%%%%%%%%%%%%%%%%%%%%%%%%%%%

\noindent \textbf{Generative methods.}
Generative methods learn features via self-reconstruction~\cite{han2019multi}. For instance, FoldingNet~\cite{yang2018foldingnet} leverages a graph-based encoder and a folding-based decoder to deform a canonical 2D grid onto the surface of a point cloud. 
L2g~\cite{liu2019l2g} uses a local-to-global auto-encoder to simultaneously learn the local and global structure of point clouds. 
In~\cite{chen2019deep}, a graph-based decoder with a learnable graph topology is used to push the codeword to preserve representative features. 
In~\cite{achlioptas2018learning} a combination of hierarchical Bayesian and generative models is trained to generate plausible point clouds. 
GraphTER~\cite{gao2020graphter} self-trains a feature encoder by reconstructing node-wise transformations from the representations of the original and transformed graphs. However, generative models are sensitive to transformations, weakening the learning of robust point cloud representations for different downstream tasks. Moreover, it is not always feasible to reconstruct back the shape from pose-invariant features \cite{sanghi2020info3d}.

\noindent \textbf{Discriminative methods.}
Discriminative methods are based on auxiliary handcrafted prediction tasks to learn point cloud representations. 
Jigsaw3D~\cite{sauder2019self} uses a 3D jigsaw puzzle approach as the self-supervised learning task. Recently, contrastive approaches~\cite{rao2020global,sanghi2020info3d,chen2021exploring, xie2020pointcontrast}, which are robust to transformation, achieved state-of-the-art performance. Info3D~\cite{sanghi2020info3d} maximizes the mutual information between the 3D shape and a geometric transformed version of the 3D shape. PointContrast~\cite{xie2020pointcontrast} is the first to research a unified framework of the contrastive paradigm for 3D representation learning.
We argue that the success of contrastive methods relies on the correct design of negative mining strategies and on the correct choice of data augmentations that should not affect the semantics of the input.
To mitigate these issues, we propose an unsupervised learning method, \ourmethod, which is formulated by implementing an EM-like algorithm and provides point-level supervision to extract discriminative point-wise features. \ourmethod needs no data augmentation procedures to learn informative representations.

%%%%%%%%%%%%%%%%%%%%%%%%%%%%%%

%%%%%%%%%%%%%%%%%%%%%%%%%%%%%%
%%%%%%%%%%%%%%%%%%%%%%%%%%%%%%%%%%%%%%%%%%%%%%%%%%%%%%%%%%%
%%%%%%%%%%%%%%%%%%%%%%%%%%%%%%%%%%%%%%%%%%%%%%%%%%%%%%%%%%%
%%%%%%%%%%%%%%%%%%%%%%%%%%%%%%%%%%%%%%%%%%%%%%%%%%%%%%%%%%%
\vspace{-.5cm}
\section{Proposed method}
\begin{figure}[t]
\vspace{-.15cm}
\centering
\footnotesize
\includegraphics[width=0.85\columnwidth]{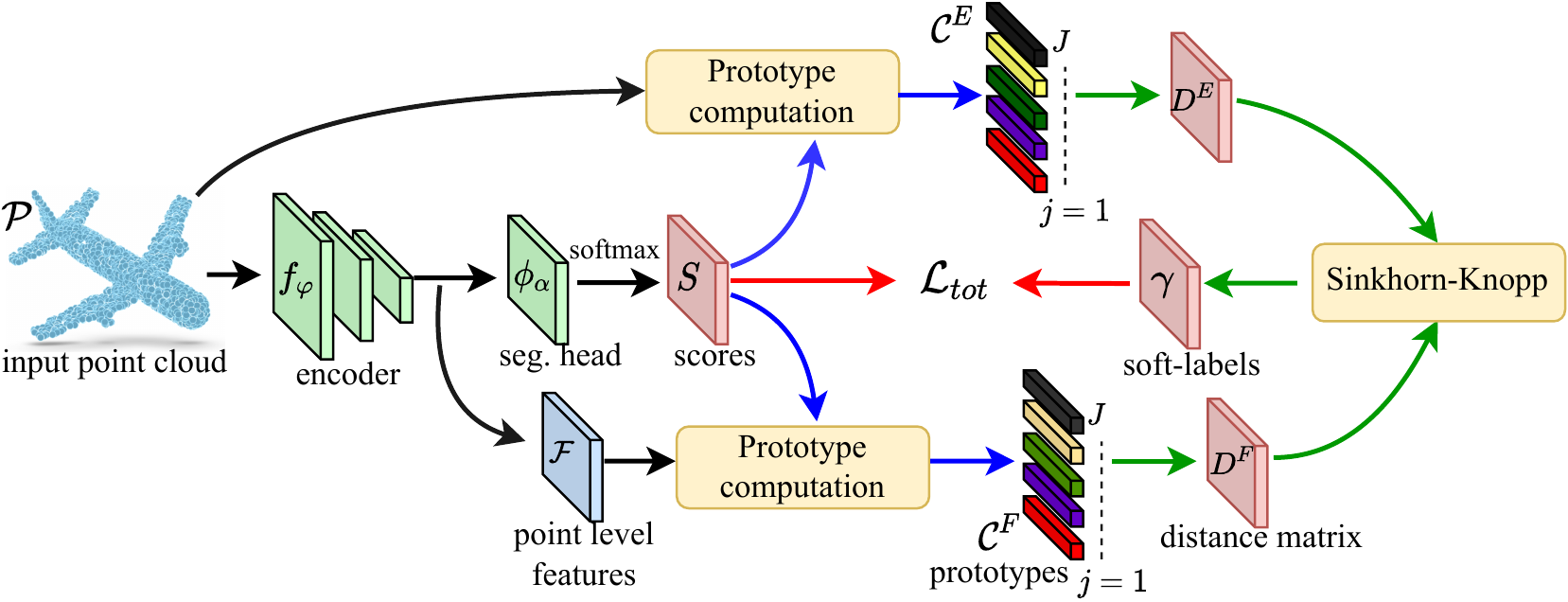}
\caption{The architecture of our \ourmethod. 
 It consists of three steps: prototype computation (blue line), soft-label $\bm{\gamma}$ assignment (green line), and optimization (red line).}\label{fig:pcl}
\end{figure}

\vspace{-.3cm}
%%%%%%%%%%%%%%%%%%%%%%%%%%%%%%%%%%%%%%%%%%%%%%%%%%%%%%%%%%%
%%%%%%%%%%%%%%%%%%%%%%%%%%%%%%%%%%%%%%%%%%%%%%%%%%%%%%%%%%%
We formulate the problem of representation learning as a soft-clustering problem.
Figure~\ref{fig:pcl} illustrates our framework that consists of three steps: prototype computation, soft-label assignment, and optimization. 
Given a 3D point cloud as an unordered set $\bm{\mathcal{P}}=\{\bm{p}_i\}_{i=1}^N$ of $N$ points where each point $\bm{p}_i \in \mathbb{R}^{3}$ is represented by a 3D coordinate $\bm{p}_i = \{x, y, z\}$, our goal is to train, in an unsupervised way, a feature encoder $f_\varphi$ with parameters $\varphi$ ($e.g.$, PointNet) that extracts informative point-wise features $\bm{\mathcal{F}} = \{f_\varphi(\bm{p}_i)\}_{i=1}^N$ from $\bm{\mathcal{P}}$. To this end, we apply a segmentation head $\phi_\alpha$ that takes as input $\bm{\mathcal{F}}$ and outputs joint log probabilities, and a softmax operator that acts on log probabilities to generate a classification score matrix $\bm{S}$. 
The prototype computation block estimates the $J$ cluster centroids (prototypes) $\bm{C}^E$ and $\bm{C}^F$ to represent each partitions. 
Next, soft-labels $\bm{\gamma}_{ij}\in \bm{\gamma}$ of each input point $\bm{p}_i$ are based on these prototypes and we use the Sinkhorn-Knopp~\cite{cuturi2013sinkhorn} algorithm to perform the soft-label assignment, i.e., $\bm{\gamma}$ softly groups $\bm{\mathcal{P}}$ into partitions. $\bm{\gamma}_{ij}\in[0,1]$ is a soft-label score that point $\bm{p}_i$ belongs to cluster $j$. The final optimization step is to minimize the average cross-entropy loss $\mathcal{L}_{tot}$ between the soft-label $\bm{\gamma}$ and the predicted class probability $\bm{S}$. 

\noindent \textbf{Prototype computation.}
We begin by computing a prototype for each cluster (partition) as the most representative feature for a set of points. 
Specifically, we use the point-wise features $\mathcal{F} = \{\bm{f}_i\}_{i=1}^N$, where $\bm{f}_i = f_\varphi(\bm{p}_i)$, to compute a set of classification scores $\bm{S}$ as $    \bm{S} = \{ \sigma(\phi_\alpha(\bm{f}_i) \}_{i=1}^N$.
% %-------------------------------------
% \begin{equation}
%     \bm{\mathcal{S}} = \{ \sigma(\phi_\alpha(\bm{f}_i) \}_{i=1}^N,
% \end{equation}
% %-------------------------------------
$\sigma$ is the softmax operation, and $\phi_\alpha$ is the segmentation layer with parameters $\alpha$. For each feature $\bm{f}_i$, $\phi_\alpha$ produces a probability score $s_{ij}$ indicating the likelihood that $\bm{p}_i$ belongs to partition $j$.
We use two types of prototypes as the representatives for each category, one in the feature space and another in the geometric space.
Specifically, according to the type, we compute $J$ prototypes as the weighted average of the features $\bm{\mathcal{F}}$ or 3D coordinates $\bm{\mathcal{P}}$ based on their scores $\bm{S}$.
Let $\bm{C}^E = \{ \bm{c}^E_j \}_{j=1}^J$ and $\bm{C}^F = \{ \bm{c}^F_j \}_{j=1}^J$ be the set of prototypes in the geometric and the features space, respectively, which are defined as
%-------------------------------------
\vspace{-.2cm}
\begin{equation} \label{eq:center_coords}
\vspace{-.2cm}
\bm{c}^E_j =\frac{1}{\sum_{i=1}^{N}s_{ij}}\sum_{i=1}^{N}s_{ij}\bm{p}_i,\quad \bm{c}^F_j =\frac{1}{\sum_{i=1}^{N}s_{ij}}\sum_{i=1}^{N}s_{ij}\bm{f}_{p_i}.
\end{equation}

%-------------------------------------
% \subsubsection{Class probability prediction}
% As shown in Fig. \ref{fig:cpp}, our model starts with a classification head $\phi_\alpha$ that takes point-wise feature $\bm{f}_{p_i}$ as input and yields a score (logit) vector $\bm{g}_{i}=\left(g_{i1},g_{i2}\cdots g_{iJ}\right)$, i.e., $\bm{g}_{i}=\phi_\alpha\left(\bm{f}_{p_i}\right)$. 

% \cris{TODO: move in implementation details
% $\phi_\alpha$ is formed by 3 fully connected layers. Each layer consists of a linear layer followed by batch normalization. Except the final layer, each layer has a LeakyReLU \cite{xu2015empirical} activation.} 

% The last layer outputs $N$ vectors with $J$ dimension, which is the same as the number of segmentation categories. The total logit predictions can be summarized by the score matrix $\bm{G}=\{g_{ij}\}_{i,j}^{N,J}$ that has size $N\times J$. The predicted class probability $s_{ij}$ of $\bm{p}_i$ that belongs to the $j$-th category is calculated by applying a row-wise softmax operation over $\bm{G}$, i.e.,
% \begin{equation}\label{eq:score}
% 	s_{ij} = \frac{\exp\left(g_{ij}\right)}{\sum_{l}^{J}\exp\left(g_{il}\right)}.
% \end{equation}

%%%%%%%%%%%%%%%%%%%%%%%%%%%%%%%%%%%%%%%%%%%%%%%%%%%%%%%%%%%%%%%
%%%%%%%%%%%%%%%%%%%%%%%%%%%%%%%%%%%%%%%%%%%%%%%%%%%%%%%%%%%%%%%
\noindent \textbf{Soft-label assignment.}
We introduce the soft-label assignment step that labels points based on their distance to the prototypes estimated by Eq.~\eqref{eq:center_coords}.
If we only use features for soft-label assignment, this would highly likely produce disconnected and scattered clusters.
Hence, we concatenate point coordinates with the features so that the label is more localized.
Specifically, we encode the pseudo-labels $\bm{\gamma}=\{\gamma_{ij}\in[0,1]\}_{i,j}^{N,J}$ as posterior distributions, i.e.~soft-labels, satisfying $\sum_{j=1}^{J}\gamma_{ij}=1$. $\gamma_{ij}$ is the posterior probability that $\bm{p}_i$ belongs to partition $j$. 
We base the assignment of soft-labels to the respective points on the prototypes $\bm{C}^E$ and $\bm{C}^F$, and by following two assumptions:
%******************
\begin{enumerate}[noitemsep,topsep=0pt,label=\roman*)]
	\item Cluster cohesion: If a point $\bm{p}_i$ belongs to partition $j$, point $\bm{p}_i$ and prototype $\bm{c}^E_{j}$ should have the shortest distance among the distances of $\bm{p}_i$ with other prototypes in $\bm{C}^E$. The same holds true in the feature space.
	\item Uniform distribution: Each point cloud is assumed to be segmented into equally-sized partitions of $\left \lfloor \frac{N}{J} \right \rfloor$ elements, where $\left \lfloor \cdot \right \rfloor$ indicates the greatest integer less than or equal to its argument.
\end{enumerate}
%******************

Assumption i) inspires us to label points based on their distance from the centroids. It can be formalized as an expression, i.e., if $\bm{p}_i$ belongs to cluster $j$, then $\|\bm{p}_i-\bm{c}^E_{j}\|_2\leq\|\bm{p}_i-\bm{c}^E_{k}\|_2, \|\bm{f}_i-\bm{c}^F_{j}\|_2\leq\|\bm{f}_i-\bm{c}^F_k\|_2$ and $\gamma_{ij}\geq \gamma_{ik}, k\neq  j, k=1,\cdots,J$. 
$\|\cdot\|_2$ is the L2 norm. This can be ensured by minimizing the following objective,
%-------------------------------------
\vspace{-.2cm}
\begin{equation}\label{eq:vgamma}
\vspace{-.2cm}
    \min_{\bm{\gamma}} \frac{1}{N}\sum_{i=1}^{N}\sum_{j=1}^{J}\left(\lambda\|\bm{p}_i-\bm{c}^E_j\|^2_2+\left(1-\lambda)\right\|\bm{f}_i-\bm{c}^F_j\|^2_2\right)\gamma_{ij},
\end{equation}
%-------------------------------------
where $\lambda \in [0,1]$ is a learned parameter. For convenience, we define the following matrix form $\bm{D}=\lambda\bm{D}^E+\left(1-\lambda\right)\bm{D}^F$, where $\bm{D}^F=\{\|\bm{f}_i-\bm{c}^F_j\|^2_2\}_{i,j}^{N,J}$ and $\bm{D}^E=\{\|\bm{p}_i-\bm{c}^E_j\|^2_2\}_{i,j}^{N,J}$ are matrices of size equal to $N\times J$. Then, Eq. (\ref{eq:vgamma}) can be rewritten as:
%-------------------------------------
\vspace{-.2cm}
\begin{equation}\label{eq:mgamma}
\vspace{-.2cm}
    \min_{\bm{\gamma}} \left<\frac{\bm{\gamma}}{N}, \bm{D}\right>,
\end{equation}
%-------------------------------------
where $\left < \cdot, \cdot \right>$ is the Frobenius matrix dot product.

Assumption ii) is formulated in a constraint condition as $\sum_{i=1}^{N}\gamma_{ij}=\frac{N}{J}$, which can mitigate the problem that all data points are assigned to a single (arbitrary) label.
Therefore, based on $\sum_{i=1}^{N}\gamma_{ij}=\frac{N}{J}$ and the property of the posterior probability $\sum_{j=1}^{J}\gamma_{ij}=1$, $\bm{\gamma}$ satisfies the following constraints 
%-------------------------------------
\vspace{-0.2cm}
\begin{equation}\label{eq:constraint}
\vspace{-.2cm}
   \frac{1}{N}\bm{\gamma}^\top\bm{1}_N  =\frac{1}{J}\bm{1}_J,
\frac{1}{N}\bm{\gamma}\bm{1}_J =\frac{1}{N}\bm{1}_N, 
\end{equation}
%-------------------------------------
where $\bm{1}_k (k=N,J)$ denotes the vector of ones in dimension $k$.

Let $\bm{\Gamma}=\frac{\bm{\gamma}}{N}$ with elements defined as $\Gamma_{ij}=\frac{\gamma_{ij}}{N}$. $\bm{\Gamma}$ satisfies $\sum_{ij}^{NJ}\bm{\Gamma}_{ij}=1$. By replacing the variable $\bm{\gamma}$ with $\bm{\Gamma}$ in Eq. (\ref{eq:mgamma}) and Eq. (\ref{eq:constraint}), the joint objective of assumptions i) and ii) can be formulated as an optimal transport (OT) problem~\cite{peyre2019computational} as
%-------------------------------------
\vspace{-0.3cm}
\begin{equation}\label{eq:poster}
    \min_{\bm{\Gamma}} \left<\bm{\Gamma}, \bm{D}\right>, ~ \mbox{s.t.} ~ \bm{\Gamma}^\top\bm{1}_N=\frac{1}{J}\bm{1}_J, \bm{\Gamma}\bm{1}_J=\frac{1}{N}\bm{1}_N.
\end{equation}
%-------------------------------------
The minimization of Eq.~(\ref{eq:poster}) can be solved in polynomial time as a linear program. However, the linear program involves millions of data points and thousands of classes and traditional algorithms hardly scale to large problems \cite{cuturi2013sinkhorn}.
We address this issue by adopting an efficient version of the Sinkhorn-Knopp algorithm~\cite{cuturi2013sinkhorn}. 
Our implementation of the Sinkhorn-Knopp algorithm is described in Alg.~\ref{alg:sinkhorn} (Appendix).
This requires the following regularization term
%-------------------------------------
\vspace{-0.3cm}
\begin{equation}\label{eq:opt}
	\min_{\bm{\Gamma}} \left<\bm{\Gamma}, \bm{D}\right> - \epsilon H\left(\bm{\Gamma}\right), 
	\quad \mbox{s.t.}~\bm{\Gamma}^\top\bm{1}_N=\frac{1}{J}\bm{1}_J, \,\,\,
		\bm{\Gamma}\bm{1}_J=\frac{1}{N}\bm{1}_N,
\end{equation}
%-------------------------------------
where $H\left(\bm{\Gamma}\right)=\left<\bm{\Gamma},\log\bm{\Gamma}-1\right>$ denotes the entropy of $\bm{\Gamma}$ and $\epsilon > 0$ is a regularization parameter. For very small $\epsilon$, optimizing Eq.~(\ref{eq:opt}) is equivalent to optimizing Eq.~(\ref{eq:poster}), but even for moderate values of $\epsilon$, the objective tends to have approximately the same optimizer \cite{cuturi2013sinkhorn}. 
The larger the $\epsilon$, the faster the convergence, please refer to~\cite{cuturi2013sinkhorn} for details. In our case, using a fixed $\epsilon=1e-3$ is appropriate as we are interested in the final clustering and representation learning results, rather than in solving the transport problem exactly. The solution to Eq.~(\ref{eq:opt}) takes the form of the following normalized exponential matrix~\cite{cuturi2013sinkhorn},
%-------------------------------------
\vspace{-.2cm}
\begin{equation}\label{eq:gamma}
\vspace{-.2cm}
	\bm{\Gamma} = \mbox{diag}\left(\bm{\mu}\right)\exp\left(\bm{D}\big/\epsilon\right)\mbox{diag}\left(\bm{\nu}\right),
\end{equation}
%-------------------------------------
where $\bm{\mu}=(\mu_1, \mu_2,\cdots,\mu_N)$ and $\bm{\nu}=(\nu_1,\nu_2,\cdots,\nu_J)$ are renormalization vectors in $\mathbb{R}^N$ and $\mathbb{R}^J$. The vectors $\bm{\mu}$ and $\bm{\nu}$ can be obtained by iterating the updates via $\bm{\mu}_i=\left[\exp\left(\bm{D}\big/\epsilon\right)\bm{\nu}\right]^{-1}_i$ and $\bm{\nu}_j=\left[\exp\left(\bm{D}\big/\epsilon\right)^\top\bm{\mu}\right]^{-1}_j$ with initial values $\bm{\mu}=\frac{1}{N}\bm{1}_N$ and $\bm{\nu}=\frac{1}{J}\bm{1}_J$, respectively. The initialization of $\bm{\mu}$ and $\bm{\nu}$ can be any distribution, and choosing the constraints as initial values allow a faster convergence~\cite{cuturi2013sinkhorn}. $\left[ \cdot\right]^{-1}_j$ defines as the inverse value of the $j^{th}$ element of its argument.
In all our experiments, we use 20 iterations as we found it works well in practice. 
After solving Eq.~(\ref{eq:gamma}), we can infer the soft-label matrix as $\bm{\gamma}=N\cdot\bm{\Gamma}$.
\noindent \textbf{Optimization.}
The optimization step follows an EM-like scheme where the Expectation step E optimizes prototypes and soft labels, while the Maximization step M optimizes the trained parameters for representation learning.
Each step can be detailed as follows:
\begin{itemize}[noitemsep,topsep=0pt]
    \item E: Given the current encoder and segmentation layer, we compute prototypes $\bm{C^E}$ and $\bm{C}^F$ following Eq.~(\ref{eq:center_coords}), and obtain soft-labels $\bm{\gamma}$ through $\bm{\gamma}=N\cdot\bm{\Gamma}$.
    \item M: Given the current soft-labels $\bm{\gamma}$ from step E, we optimize the encoder $f_\varphi$ and  segmentation layer $\phi_\alpha$ parameters.
\end{itemize}
During E, we solve the OT problem with the Sinkhorn-Knopp algorithm. 
During M, we minimize the segmentation loss based on the resulting soft labels, such as
%-------------------------------------
\vspace{-.3cm}
\begin{equation}\label{eq:sup}
\vspace{-.2cm}
    \mathcal{L}_{soft}(\bm{\gamma}, \bm{S}) = -\frac{1}{N}\left<\bm{\gamma},\log\bm{S}\right>=-\frac{1}{N}\sum_{i=1}^{N}\sum_{j=1}^{J}\bm{\gamma}_{ij}\log s_{ij},
\end{equation}
%-------------------------------------
which corresponds to the minimization of the standard cross-entropy loss between soft-labels $\bm{\gamma}$ and predictions $\bm{S}$. 
However, the optimization of Eq.~(\ref{eq:sup}) does not ensure encoder $f_\varphi$ from predicting the same features for all the points, i.e.~all centroids collapse into the same vector.
We further promote centroids separation by minimizing the orthogonal regularization loss $\mathcal{L}_{orth}(\bm{C}) = \|{\bm{C}^E}_*^\top\bm{C}^E_*-\bm{I}\|_{Fr} + \|{\bm{C}^F}_*^\top\bm{C}^F_*-\bm{I}\|_{Fr}$,
% %-------------------------------------
% \begin{equation}\label{eq:orth}
% 	\mathcal{L}_{orth}(\bm{C}) = \|{\bm{C}^E}_*^\top\bm{C}^E_*-\bm{I}\|_{Fr} + \|{\bm{C}^F}_*^\top\bm{C}^F_*-\bm{I}\|_{Fr},
% \end{equation}
% %-------------------------------------
where $\bm{C}^k_*=[\frac{\bm{c}^k_1}{\|\bm{c}^k_1\|_2},\frac{\bm{c}^k_2}{\|\bm{c}^k_2\|_2},\cdots,
\frac{\bm{c}^k_J}{\|\bm{c}^k_J\|_2}]$ with $k=E, F$, and $\|\cdot\|_{Fr}$ is Frobenius norm. 
We define the final objective loss for the M step as
%-------------------------------------
\vspace{-.3cm}
\begin{equation}
\vspace{-.2cm}
    \mathcal{L}_{tot} = \mathcal{L}_{soft} + \eta \mathcal{L}_{orth},
\end{equation}
%-------------------------------------
where $\eta=0.01$ is a weighting parameter. We set the value of $\eta$ empirically and found that $\eta\leq 0.01$ can slightly improve the performance. 
The minimization of this loss leads to the maximization of the expected number of points correctly classified, associating the correct neighbor prototypes.
This facilitates the encoder to learn more local geometric information. 
Our implementation of \ourmethod is described in Alg.~\ref{alg:cluster} in the supplementary material.
\vspace{-.6cm}
\section{Experiments}\label{doc:exp}
\vspace{-.2cm}
In this section, we present the implementation details, the setup of pre-training and the downstream fine-tuning.
We invite the reader to check in the supplementary material for additional results on few-shot learning, ablation studies, Transformer, and additional visualizations.

%%%%%%%%%%%%%%%%%%%%%%%%%%%%%%%%%%%%%%%%%%%%%%%%%%%%%%%%%%%
\vspace{-.4cm}
\subsection{Pre-training setup}
We explore pre-training strategies on single objects (ShapeNet~\cite{shapenet2015}) and complex scenes with multiple objects (ScanNet~\cite{dai2017scannet}) to evaluate the effectiveness of \ourmethod. We implemented \ourmethod in PyTorch and executed our experiments on two Tesla V100-PCI-E-32G GPUs. During pre-training, we set $J=64$ and $\epsilon=1e-3$.

%%%%%%%%%%%%%%%%%%%%%%%%%%%%%%%%%%%%%%%%%%%%%%%%%%%%%%%%%%%
\noindent\textbf{ShapeNet}~\cite{shapenet2015} is a collection of single-object CAD models and contains $57448$ synthetic objects from 55 object categories.
We follow the experimental setup presented in \cite{sauder2019self,huang2021spatio}, we use PointNet~\cite{qi2017pointnet} and DGCNN~\cite{wang2019dynamic} as encoder networks.
The latent dimension of both encoders is $1024$. 
Following~\cite{huang2021spatio}, each point cloud is randomly downsampled to $2048$ points.
We use the official training split of ShapeNet for pre-training. 
Our pre-training involves $250$ epochs by using the AdamW \cite{gugger2018adamw} optimizer, the batch size is equal to $32$, and initial learning is equal to $0.001$ that decays by $0.7$ every $20$ epochs.

\noindent\textbf{ScanNet}~\cite{dai2017scannet} is a dataset of indoor scenes with multiple objects and consists of 1513 reconstructed meshes for 707 unique scenes. Following~\cite{xie2020pointcontrast}, we choose SR-UNet provided in PointContrast~\cite{xie2020pointcontrast} as backbone.
For pre-training, we use an SGD optimizer with a learning rate of 0.1 and a batch size of 32.  The learning rate is decreased by a factor of 0.99 every 1K iteration.
The model is trained for 30K iterations.

\vspace{-0.4cm}
\subsection{Downstream fine-tuning}\label{sec:evaluation}
\vspace{-0.2cm}
We evaluate \ourmethod on three downstream tasks: classification, part segmentation, and semantic segmentation.
We compare \ourmethod to state-of-the-art discriminative approaches (Jigsaw3D~\cite{sauder2019self}, STRL~\cite{huang2021spatio}, CrossPoint~\cite{afham2022crosspoint}, SimCLR~\cite{chen2020simple}, STRL~\cite{huang2021spatio}, PointContrast~\cite{xie2020pointcontrast}, and ContrastiveScene~\cite{hou2021exploring}) and a generative approach (OcCo~\cite{wang2020unsupervised} and ParAE~\cite{eckart2021self}).
The details of setups for downstream tasks can be found in supplementary for three encoders.
%++++++++++++++++++++++++++++++++++++++

\begin{figure}[t]
	\centering
	\begin{minipage}{.48\textwidth}
		\centering
		\begin{tabular}{cc}
			\includegraphics[width=.48\textwidth]{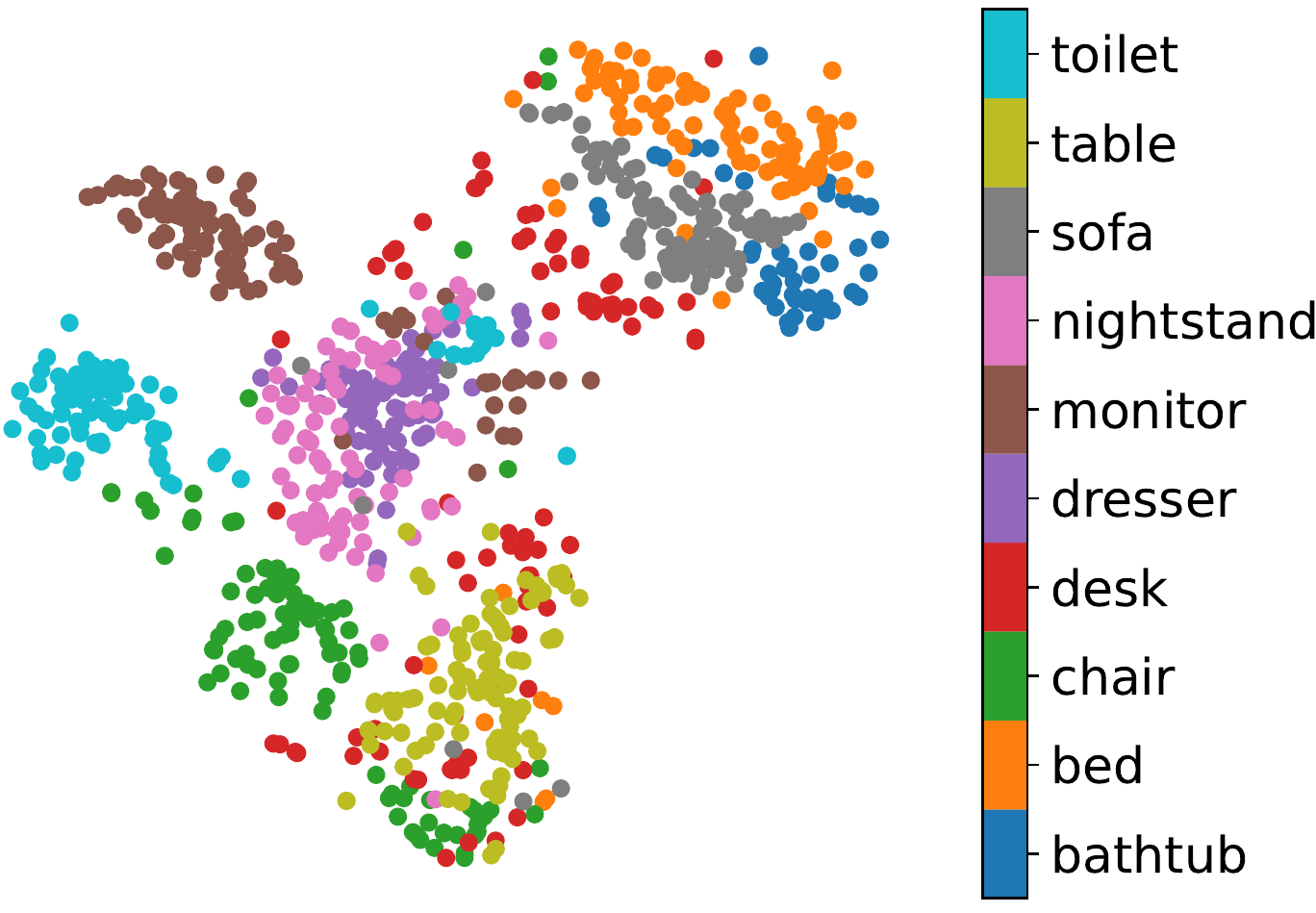} &
			\includegraphics[width=.48\textwidth]{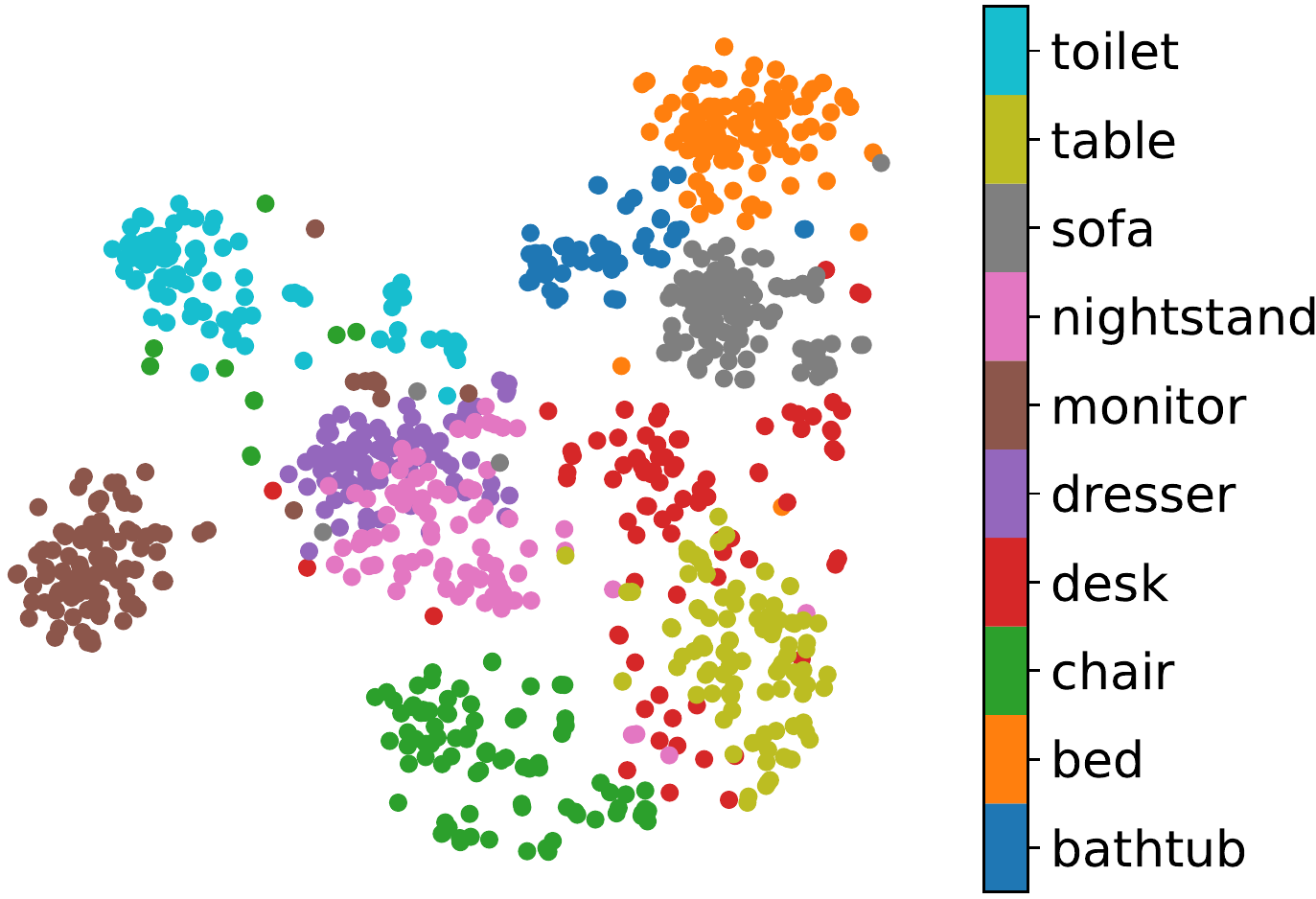}
		\end{tabular}
		\vspace{-.15cm}
		\footnotesize
		\caption{T-SNE-processed network features obtained with (left) OcCo~\cite{wang2020unsupervised} and (right) \ourmethod on ModelNet10.}
		\label{fig:tsne}
	\end{minipage}
	\hspace{.2cm}
	\begin{minipage}{.48\textwidth}
		\centering
		\begin{tabular}{cc}
			\includegraphics[width=.41\textwidth]{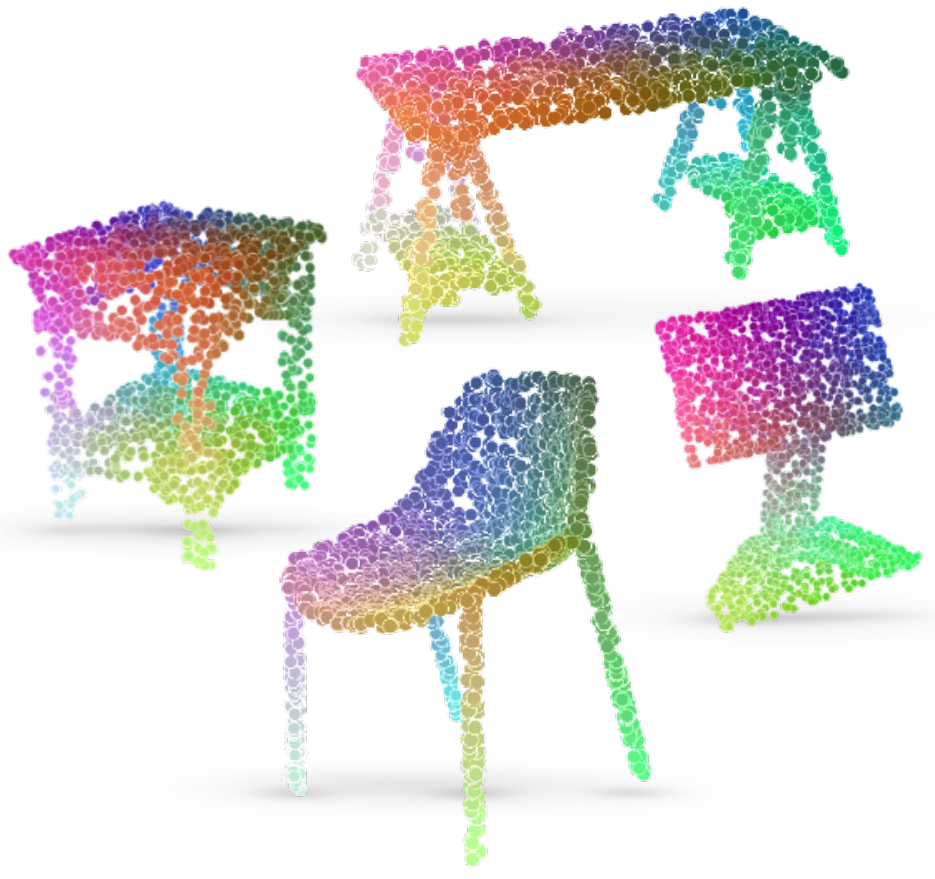} &
			\includegraphics[width=.41\textwidth]{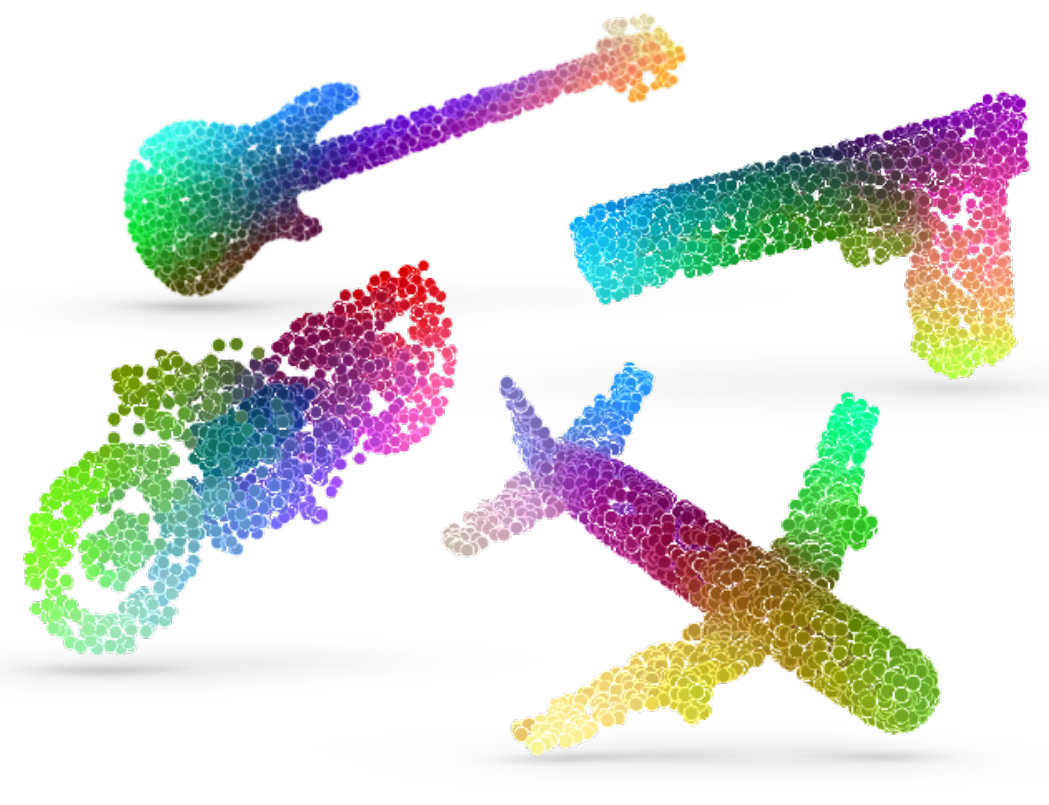}
		\end{tabular}
		\vspace{-.15cm}
		\footnotesize
		\caption{Color-coded points based on PCA projections of the learned features: (left) ModelNet40, (right) ShapeNet.}
		\label{fig:pca}
	\end{minipage}
	\vspace{-0.4cm}
\end{figure}

%%%%%%%%%%%%%%%%%%%%%%%%%%%%%%%%%%%%%%%%%%%%%%%%%%%%%%%%%%%
\noindent \textbf{Object classification.}
We use linear SVM classification on ModelNet40~\cite{sharma2016vconv} and ModelNet10~\cite{sharma2016vconv} datasets to evaluate the quality of their pre-trained versions on ShapeNet. 
Table~\ref{tab:classification} reports the classification accuracy of \ourmethod, compared to the other approaches. 
Results show that the \ourmethod is more effective than the alternative pre-training methods on both datasets.
Specifically, on ModelNet40, \ourmethod with PointNet backbone achieves the same classification accuracy ($90.3\%$) as ParAE~\cite{eckart2021self} while outperforming the contrastive approach STRL \cite{huang2021spatio} ($88.3\%$). 
The linear SVM classification performance of our method even surpasses the performance of the fully supervised PointNet, which achieves an 89.2\% test accuracy.
With the DGCNN encoder, our method achieves a $91.9\%$ test accuracy, outperforming ParAE ($91.6\%$) by $0.3\%$, and generating model OcCo~\cite{wang2020unsupervised} by $2.7\%$. 
On ModelNet10, \ourmethod outperforms OcCo~\cite{wang2020unsupervised} and Jigsaw3D~\cite{sauder2019self} with both the encoding networks. Compared to jigsaw tasks that coarsely segment a point cloud into disjoint partitions, \ourmethod learns the partitioning function itself to softly assign point clouds into coherent clusters.
Moreover, we also report results (SoftClu$^\star$) of \ourmethod pre-trained on ModelNet40, since some methods are pretrained on ModelNet40.
\ourmethod also achieves competitive results. 
We further use visualization to explore pre-trained features before fine-tuning with the DGCNN encoder. 
Figure~\ref{fig:tsne} shows the features visualized with T-SNE~\cite{van2008visualizing} of OcCo and \ourmethod on ModelNet10. 
Our method yields a better separation of the features than OcCo, which indicates a better ability of \ourmethod in clustering objects in the feature space. 
In Figure~\ref{fig:pca}, we color the points based on the PCA projections of the network features and shows how the pre-trained encoder can effectively embed geometric information.

%++++++++++++++++++++++++++++++++++++++
\begin{table}[t]
	\centering
	\footnotesize
	\caption{Linear classification comparisons on ModelNet40 and ModelNet10. $\star$ indicates that models are pre-trained on ModelNet40, otherwise, models are pre-trained on ShapeNet.}
	\label{tab:classification}
	\resizebox{0.78\columnwidth}{!}{%
		\begin{tabular}{l c c c c c}
			\toprule
			\multirow{2}{*}{Method} &
			\multirow{2}{*}{Year} &
			\multicolumn{2}{c}{PointNet} & \multicolumn{2}{c}{DGCNN}\\
			\cline{3-4} \cline{5-6}
			\multicolumn{2}{c}{} & ModelNet40 & ModelNet10 & ModelNet40 & ModelNet10 \\
			\toprule
			DeepCluster~\cite{caron2018deep} & 2018 & 86.3 & 91.6 & 90.4 & 94.1 \\
			Jigsaw3D$^\star$~\cite{sauder2019self} & 2019 & 87.5 & 91.3 & 87.8 & 92.6 \\
			Jigsaw3D~\cite{sauder2019self} & 2019 & 87.3 & 91.6 & 90.6 & 94.5 \\
			Rotation3D~\cite{poursaeed2020self} & 2020 & 88.6 & - & 90.8 & -\\
			SwAV~\cite{caron2020unsupervised} & 2020 & 85.4 & 92.1 & 90.3 & 93.5 \\
			OcCo~\cite{wang2020unsupervised} & 2021 & 88.7 & 91.4 & 89.2 & 92.7 \\
			SimCLR~\cite{chen2020simple} & 2021 & 88.4 & 91.4 & 90.1 & 92.1 \\ 
			STRL~\cite{huang2021spatio} & 2021 & 88.3 & - & 90.9 & - \\
			ParAE~\cite{eckart2021self} & 2021 & \textbf{90.3} & - & 91.6 & - \\
			CrossPoint~\cite{afham2022crosspoint} & 2022 & 89.1 & - & 91.2 & - \\
			SoftClu$^\star$ (Ours) & - & 88.4 & 93.0 & 91.4 & 94.5 \\
			SoftClu (Ours) & - & \textbf{90.3} & \textbf{93.5} & \textbf{91.9} & \textbf{94.8} \\
			\bottomrule
		\end{tabular}
	}
	\vspace{-.2cm}
\end{table}

%++++++++++++++++++++++++++++++++++++++
%++++++++++++++++++++++++++++++++++++++

%%%%%%%%%%%%%%%%%%%%%%%%%%%%%%%%%%%%%%%%%%%%%%%%%%%%%%%%%%%
\noindent \textbf{Part segmentation.}
We follow~\cite{wang2020unsupervised} and use the ShapeNetPart~\cite{yi2016scalable} benchmark dataset. 
Table~\ref{tab:part_segmentation} reports the part segmentation results of \ourmethod in comparison with the alternative approaches on ShapeNetPart~\cite{yi2016scalable}.
\ourmethod outperforms all the other approaches with both PointNet and DGCNN encoders in terms of both OA and mIoU. 
With the PointNet encoder, \ourmethod achieves $93.9\%$ OA and $83.8\%$ mIoU, improving over state-of-the-art CrossPoint ($93.2\%$ OA, $82.7\%$ mIoU) by $0.7\%$ OA and $1.1\%$ mIoU. 
With the DGCNN encoder, we achieve $94.6\%$ OA and $85.7\%$ mIoU, outperforming CrossPoint ($94.4$ OA, $85.3$\% mIoU) of about $0.2\%$ OA and $0.4\%$ mIoU. 
Figure~\ref{fig:part_segmentation} in the supplementary material shows examples of qualitative part segmentation results.

%++++++++++++++++++++++++++++++++++++++
%===========================================
\begin{table}[t]
	\vspace{-0.2cm}
	\begin{minipage}{.49\textwidth}
		\centering
		\footnotesize
		\caption{Part segmentation results.}
		\label{tab:part_segmentation}
		\resizebox{.90\textwidth}{!}{
			\begin{tabular}{clcccc}
				\toprule
				\multirow{2}{*}{Encoder} & 
				\multirow{2}{*}{Method} &
				\multicolumn{2}{c}{Metrics} \\
				\cline{3-4} 
				\multicolumn{2}{c}{} & OA ($\%$) & mIoU ($\%$) \\
				\toprule
				\multirow{5}{*}{PointNet} 
				& Random & 92.8 & 82.2 \\
				& Jigsaw3D \cite{sauder2019self} & 93.1 & 82.2 \\
				& OcCo \cite{wang2020unsupervised}  & 93.4 & 83.4 \\
				& CrossPoint \cite{afham2022crosspoint}  & 93.2 & 82.7 \\
				& SoftClu (Ours) & \textbf{93.9} & \textbf{83.8} \\
				\midrule
				\multirow{5}{*}{DGCNN} 
				& Random & 92.2 & 84.4 \\
				& Jigsaw3D \cite{sauder2019self} & 92.7 & 84.3 \\
				& OcCo \cite{wang2020unsupervised}  & 94.4 & 85.0 \\
				& CrossPoint \cite{afham2022crosspoint}  & 94.4 & 85.3 \\
				& SoftClu (Ours) & \textbf{94.6} & \textbf{85.7} \\
				\bottomrule
			\end{tabular}
		}
		\vspace{-0.21cm}
	\end{minipage}
	\hspace{.2cm}
	\begin{minipage}{0.49\textwidth} 
		\centering
		\footnotesize
		\caption{Semantic segmentation results.}
		\label{tab:semantic_segmentation}
		\resizebox{.90\textwidth}{!}{
			\begin{tabular}{clcc}
				\toprule
				\multirow{2}{*}{Encoder} & 
				\multirow{2}{*}{Method} &
				\multicolumn{2}{c}{Metrics} \\
				\cline{3-4} 
				\multicolumn{2}{c}{} & OA ($\%$) & mIoU ($\%$) \\
				\toprule
				\multirow{5}{*}{PointNet}  
				& Random & 78.9 & 47.0 \\
				& Jigsaw3D \cite{sauder2019self} & 80.1 & 52.6 \\
				& OcCo \cite{wang2020unsupervised}  & 82.0 & 54.9 \\
				& CrossPoint \cite{afham2022crosspoint}  & 81.8 & 54.5 \\
				& SoftClu (Ours) & \textbf{82.9} & \textbf{55.3} \\
				\midrule
				\multirow{5}{*}{DGCNN} 
				& Random & 83.7 & 54.9 \\
				& Jigsaw3D \cite{sauder2019self} & 84.1 & 55.6 \\
				& OcCo \cite{wang2020unsupervised}  & 84.6 & 58.0  \\
				& CrossPoint \cite{afham2022crosspoint}  & 84.7 & 58.4 \\
				& SoftClu (Ours) & \textbf{85.4} & \textbf{59.2} \\
				\bottomrule
			\end{tabular}
		}
		\vspace{-0.2cm}
	\end{minipage} 
	\vspace{-0.65cm}
\end{table}
% %++++++++++++++++++++++++++++++++++++++

\noindent \textbf{Semantic segmentation.} 
We first evaluate \ourmethod features on semantic segmentation by using the S3DIS~\cite{armeni20163d} benchmark dataset, where features are pre-trained on ShapeNet. 
Table~\ref{tab:semantic_segmentation} reports the segmentation results of \ourmethod and that of the other baselines on S3DIS~\cite{armeni20163d}. 
\ourmethod outperforms all the other approaches with both PointNet and DGCNN encoders. 
With the PointNet encoder, \ourmethod achieves $82.9\%$ OA and $55.3\%$ mIoU, outperforming both the state-of-the-art OcCo ($82.0\%$ OA, $55.3\%$ mIoU) and Jigsaw3D ($80.1\%$ OA, $52.6\%$ mIoU). 
With the DGCNN encoder, \ourmethod achieves $85.4\%$ OA and $59.2\%$ mIoU, outperforming CrossPoint~\cite{afham2022crosspoint} (84.7\% OA and 58.4\% mIoU), OcCo ($84.6\%$ OA, $58.0\%$ mIoU) and Jigsaw3D ($84.1\%$ OA, $55.6\%$ mIoU).
We also compare \ourmethod with point-level methods such as PointContrast~\cite{xie2020pointcontrast} and ContrastiveScene~\cite{hou2021exploring} when features are pre-trained on ScanNet with SR-UNet backbone.
Table~\ref{tab:point_level} shows that our pre-training method achieves 73.4\% mIoU and 79.1\% mAcc, outperforming the pre-training results of PointContrast~\cite{xie2020pointcontrast} and ContrastiveScene~\cite{hou2021exploring}.

%++++++++++++++++++++++++++++++++++++++
\begin{table}[!hbt]
	\centering
	\vspace{-.12cm}
	\footnotesize
	\caption{Results of semantic segmentation with SR-UNet backbone~\cite{armeni20163d}.}
	\label{tab:point_level}
	\resizebox{.7\textwidth}{!}{
		\begin{tabular}{c|cccc}
			\toprule
			Method   & Scratch & PointContrast~\cite{xie2020pointcontrast} & ContrastiveScene~\cite{hou2021exploring} & \ourmethod \\
			\toprule
			mIoU   & 68.2 & 70.3 & 72.2 & \bf73.4 \\
			mAcc   & 75.5 & 76.9 & - & \bf79.1 \\
			\bottomrule
		\end{tabular}
	}
	\vspace{-0.2cm}
\end{table}
%%%%%%%%%%%%%%%%%%%%%%%%%%%%%%

%%%%%%%%%%%%%%%%%%%%%%%%%%%%%%
%%%%%%%%%%%%%%%%%%%%%%%%%%%%%%%%%%%%%%%%%%%%%%%%%%%%%%%%%%%
%%%%%%%%%%%%%%%%%%%%%%%%%%%%%%%%%%%%%%%%%%%%%%%%%%%%%%%%%%%
\vspace{-.3cm}
\subsection{Ablation study and analysis} \label{sec:ablations}
\vspace{-.2cm}
%++++++++++++++++++++++++++++++++++++++
%++++++++++++++++++++++++++++++++++++++
% \begin{figure}[t]
% 	\centering
% 	\includegraphics[width=1.0\columnwidth]{figures/clus_J}
% 	\caption{\cris{Results of \ourmethod on ModelNet10 with different number of clusters $J$.}}
% 	\label{fig:clus_J}
% \end{figure}
%++++++++++++++++++++++++++++++++++++++
%++++++++++++++++++++++++++++++++++++++

% We first explore the influence of the number of clusters or partitions $J$ used in \ourmethod, and Table \ref{tab:proto} shows the results.  It is observed that varying the number of clusters by order of magnitude (16-128) affects the performance slightly (at most 0.5 and 0.3 on NodelNet40 for PointNet and DGCNN, respectively) when they are between 32 and 128. Table \ref{tab:proto} demonstrates that the number of clusters has little influence, as long as they are ``enough". Throughout the paper,  we train \ourmethod with 64 clusters for PointNet and DGCNN as it produces a good performance. 

%%%%%%%%%%%%%%%%%%%%%%%%%%%%%%%%%%%%%%%%%%%%%%%%%%%%%%%%%%%
\noindent \textbf{Feature and geometric prototypes.}
We study the contribution of feature and geometric prototypes of \ourmethod (Eq.~\ref{eq:center_coords}).
We perform pre-training by using (i) only feature prototypes, (ii) only geometric prototypes, and (iii) both prototypes.
We perform this study on ModelNet40 and ModelNet10. 
In Table~\ref{tab:ab_space} we can observe that when only feature prototypes are used, \ourmethod achieves the lowest performance.
This is due to wrong cluster assignments of those points sharing similar features but belonging to different geometric regions (e.g.~wings of an airplane).
Using both feature and geometric prototypes, \ourmethod can achieve the best performance on ModelNet40 and ModelNet10, with both PointNet and DGCNN.
%++++++++++++++++++++++++++++++++++++++
%++++++++++++++++++++++++++++++++++++++
\begin{table}[!hbt]
	\centering
 \vspace{-.2cm}
    \footnotesize
	\caption{Ablation study of \ourmethod by using different prototypes.}
	\label{tab:ab_space}
    \resizebox{.56\textwidth}{!}{
	\begin{tabular}{l c c c c}
		\toprule
		\multirow{2}{*}{Encoder} & 
		\multirow{2}{*}{Geometry} &
		\multirow{2}{*}{Feature} &
		\multicolumn{2}{c}{Accuracy} \\
		\cline{4-5} 
		\multicolumn{2}{c}{} & & ModelNet40 & ModelNet10 \\
		\toprule
		\multirow{3}{*}{PointNet} 
		& $\checkmark$ &  & 88.7 & 92.9 \\ 
		& &  $\checkmark$ & 86.5 & 92.7 \\
		& $\checkmark$ & $\checkmark$ & \bf90.3 & \bf93.5 \\
		\midrule
		\multirow{3}{*}{DGCNN} 
		& $\checkmark$ &  & 91.4 & 94.5 \\
		& & $\checkmark$ & 90.5 & 93.3 \\
		& $\checkmark$ & $\checkmark$ & \bf91.9 & \bf94.8 \\
		\bottomrule
	\end{tabular}
	}
 \vspace{-0.2cm}
\end{table}
%++++++++++++++++++++++++++++++++++++++
%++++++++++++++++++++++++++++++++++++++

%%%%%%%%%%%%%%%%%%%%%%%%%%%%%%%%%%%%%%%%%%%%%%%%%%%%%%%%%%%
\noindent \textbf{Number of clusters.}
We assess the effect of selecting different numbers of cluster partitions $J$ by using ModelNet40.
We pre-train \ourmethod with different values of $J$, i.e.~from $16$ to $128$, and report the results in Table~\ref{tab:ablation_j}.
\ourmethod achieves the best results with $J=64$ for both PointNet and DGCNN.
We observed stability with the results throughout different values of $J$. 
Additional results of the batch size and soft-labels please refer to the supplementary material.
%++++++++++++++++++++++++++++++++++++++
\begin{table}[!hbt]
\vspace{-.2cm}
	\centering
	\caption{Ablation study results of \ourmethod with different number of clusters $J$.}
	\label{tab:ablation_j}
    \resizebox{.65\textwidth}{!}{
	\begin{tabular}{l c c c c c c c c c}
		\toprule
		Method   & 16   & 32   & 48   & 64   & 72 & 96 & 112 & 128 \\
		\toprule
		PointNet & 92.4 & 93.0 & 93.1 & 93.5 & 93.4 & 93.3 & 93.2 & 93.1 \\
		DGCNN    & 94.2 & 94.8 & 94.6 & 94.8 & 94.7 & 94.6 & 94.6 & 94.5 \\
		\bottomrule
	\end{tabular}
	}
 \vspace{-0.2cm}
\end{table}

\noindent \textbf{Running times.}
Our method is used only for pre-training, where each iteration consists of two parts: a backbone forward pass and SoftClu optimization. 
We run our method on one Tesla V100 GPU (32G) and two Intel(R) 6226 CPUs and measured the iteration time over several iterations. 
SoftClu adds an average overhead of 0.014ms for each iteration on ShapeNet with a DGCNN backbone.
Notice that, the inference time of each backbone used remains the original one.

%%%%%%%%%%%%%%%%%%%%%%%%%%%%%%

%%%%%%%%%%%%%%%%%%%%%%%%%%%%%%
%%%%%%%%%%%%%%%%%%%%%%%%%%%%%%%%%%%%%%%%%%%%%%%%%%%%%%%%%%
\vspace{-.6cm}
\section{Conclusions}
\vspace{-.2cm}
We presented SoftClu, a data augmentation-free unsupervised representation learning scheme for 3D point cloud understanding. SoftClu implicitly alternates between clustering the point-level features to produce point-wise pseudo-labels and utilizing these soft-labels to train the representations. 
Our approach showed promising results in transferring the pre-trained representations to different downstream 3D understanding tasks, such as classification, part segmentation, and semantic segmentation.
Our SoftClu is independent from specific deep network architectures, enabling us to use our method as a generic method for feature extraction from raw point cloud data to improve other 3D models performances.

\noindent \textbf{Acknowledgments.}
This work was supported by the EU H2020 AI4Media No. 951911 project and the EUREGIO project OLIVER. 
%%%%%%%%%%%%%%%%%%%%%%%%%%%%%%
%-------------------------------------------------------------------------

\bibliography{egbib}

\newpage
\newpage
\appendix
%+++++++++++++++++++++++++++++++++++++++
\begin{center}
\textbf{\Large Supplementary Materials for Data Augmentation-free Unsupervised Learning for 3D Point Cloud Understanding}
\end{center}

\section{Algorithm}
Here, we provide a pseudo-code for \ourmethod training loop in Algorithm~\ref{alg:cluster}.
\begin{algorithm}[!hbt]
\small
	\caption{Soft clustering (pseudocode).}  
	\label{alg:cluster} 
	\hspace*{\algorithmicindent} \textbf{Input:} $\{\mathcal{P}\}$ a set of 3D point clouds and each point cloud has $N$ points; $K$ number of optimization steps. \\
	\hspace*{\algorithmicindent} {\bf Output:} the backbone $f_\varphi$ pretrained by using our algorithm.
	\begin{algorithmic}[1]
		\For{$i$ in range(0, K)}
		\State $\mathcal{L}_{tot}=0$
		\For{$\mathcal{P}\in \left\{\mathcal{P}\right\}$}
		\State {\# compute class scores}
		\State $\bm{S} = \mbox{softmax}\left(\phi_\alpha\left(f_\varphi\left(\mathcal{P}\right)\right)\right)$ 
		\State {\# compute prototypes}
		\State $\bm{C}^E = \left\{\frac{1}{\sum_{i=1}^{N}s_{ij}}  \sum_{i=1}^{N}s_{ij}\bm{p}_i \right\}_{j=1}^N$
		\State $\bm{C}^F = \left\{ \frac{1}{\sum_{i=1}^{N}s_{ij}} \sum_{i=1}^{N}s_{ij}\bm{f}_i \right\}_{j=1}^N$
		\State {\# compute $\bm{D}$}
		\State $\bm{D} = \left\{\lambda\|\bm{p}_i-\bm{c}^E_j\|^2_2+\left(1-\lambda)\right\|\bm{f}_i-\bm{c}^F_j\|^2_2 \right\}_{i,j}^{N,J}$
		\State {\# compute $\gamma$}
		\State $\bm{\Gamma} = \mbox{SINKHORN}\left(\mbox{stopgrad}\left(\bm{D}\right), 1e-3, 20\right)$
		\State $\bm{\gamma}=N\cdot\bm{\Gamma}$
		\State {\# compute loss}
		\State $\mathcal{L}_{tot} \mathrel{{+}{=}} \mathcal{L}_{soft}$ + $\eta\mathcal{L}_{orth}$
		\EndFor
		\State {\# update backbone and segmentation head}
		\State $f_\varphi, \phi_\alpha \leftarrow\mbox{optimize}\left(\frac{\mathcal{L}_{tot}}{N}\right)$ 
		\EndFor \\
		\Return $f_\varphi$
	\end{algorithmic}
\end{algorithm}

For the Sinkhorn-Knopp algorithm, we provide a detailed pseudo-code in Algorithm~\ref{alg:sinkhorn}.
\begin{algorithm}[!hbt]
\small
  	\caption{Sinkhorn-Knopp algorithm (pseudocode).}  \label{alg:sinkhorn} 
	\hspace*{0.02in} {\bf Input:} {$D$ distance matrix, $\epsilon=1e-3$ and $niters$ iterations.} 
  \begin{algorithmic}[1]
    \Function{sinkhorn}{$\bm{D}$, $\epsilon$, $niters$}
    	\State $\bm{\Gamma} = \exp(\bm{D} / \epsilon)$
		\State $\bm{\Gamma} /= \mbox{sum}(\bm{\Gamma})$  
		\State $N, J = \bm{\Gamma}$.shape
		\State $\bm{u}, \bm{\mu}, \bm{\nu} = \mbox{zeros}(N),\mbox{ones}(N) / N, \mbox{ones}(J) / J$ 
		\For{ \_ in range(0, \textit{niters})}
		\State $\bm{u} = $ sum($\bm{\Gamma}$, dim=1)
		\State $\bm{\Gamma} *= (\bm{\mu} / \bm{u})$.unsqueeze(1) 
		\State $\bm{\Gamma} *= (\bm{\nu} / \mbox{sum}(\bm{\Gamma}, \mbox{dim=0}))$.unsqueeze(0)
		\EndFor\\
		\Return $\bm{\Gamma}$
    \EndFunction
  \end{algorithmic}
\end{algorithm}

\section{Downstream Tasks Setups}
%%%%%%%%%%%%%%%%%%%%%%%%%%%%%%%%%%%%%%%%%%%%%%%%%%%%%%%%%%%
\noindent \textbf{Classification.} 
We use ModelNet40~\cite{sharma2016vconv} and ModelNet10~\cite{sharma2016vconv} benchmark classification datasets. ModelNet40 is composed of $12331$ meshed models from $40$ object categories, split into $9843$ training meshes and $2468$ testing meshes, where the points are sampled from CAD models.
ModelNet10 dataset contains $4899$ pre-aligned shapes from $10$ categories with $3991$ ($80\%$) shapes for training and $908$ ($20\%$) shapes for testing. 
For SVM training, we randomly sample $1024$ points for each shape as in \cite{sauder2019self}.

\noindent \textbf{Part segmentation.} 
We follow~\cite{wang2020unsupervised} and use the ShapeNetPart~\cite{yi2016scalable} benchmark dataset that contains $16881$ objects of $2048$ points from $16$ categories with $50$ parts in total. 
We train the linear fully connected layer for $100$ epochs by using the AdamW \cite{gugger2018adamw} optimizer with batch size of $24$, initial learning rate of $0.001$, learning rate decay of $0.5$ every $20$ epoch. 
We report the overall accuracy (OA) and the mean class intersection over union (mIoU) to evaluate segmentation quality as in \cite{wang2020unsupervised}.

\noindent \textbf{Semantic segmentation.} 
We evaluate \ourmethod features on semantic segmentation by using the S3DIS~\cite{armeni20163d} benchmark dataset. 
S3DIS consists of 3D scans collected with the Matterport scanner in six indoor areas, featuring 271 rooms and 13 semantic classes. 
Following the pre-processing, post-processing and training settings as in \cite{wang2020unsupervised}, we split each room into $1m \times 1m$ blocks and use 4,096 points as the model input. For PointNet and DGCNN, we finetune the pre-trained model on areas 1,2,3,4,6 and test them on area 5. As in part segmentation, we report OA and mIoU to quantify the segmentation quality.

For SR-UNet backbone, we finetune the pre-trained model on areas 1-4 and 6 and test them on area 5. The fine-tuning experiments are trained with a batch size of 48 for a total of 10K iterations. 
The initial learning rate is 0.1, with polynomial decay with a power of 0.9. We set voxel size as 0.05 (5cm) and weight decay as 0.0001.
We report mIoU and mAcc to evaluate segmentation quality as in \cite{xie2020pointcontrast}.

\section{More Results}
\begin{figure}[t]
	\centering
 \vspace{-.3cm}
	\includegraphics[width=0.98\textwidth]{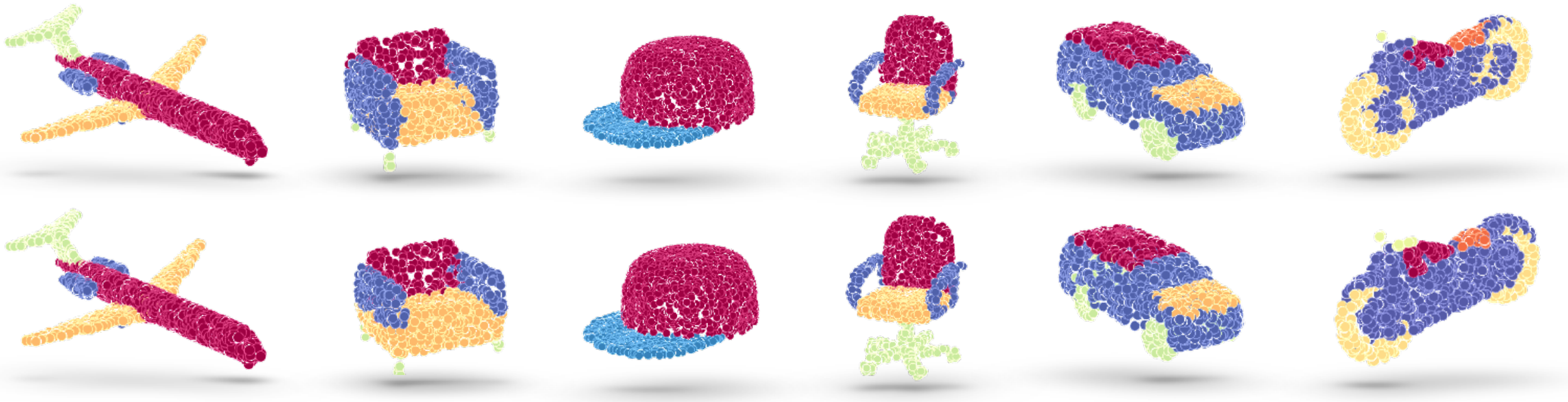}
	\put(-503,90){\color{black}\footnotesize \rotatebox{90}{\textbf{Ours}}}
	\put(-503,30){\color{black}\footnotesize \rotatebox{90}{\textbf{GT}}}
	\caption{Part segmentation results on ShapeNetPart \cite{yi2016scalable} of \ourmethod using the DGCNN encoder (top row) compared to the ground-truth annotations (bottom row).}
	\label{fig:part_segmentation}
\end{figure}

\paragraph{Part segmentation visualizations.} Figure~\ref{fig:part_segmentation} shows examples of qualitative part segmentation results obtained with \ourmethod after the fine-tuning on the downstream task compared to ground-truth annotations (GT).
We can observe that our method provides consistent predictions throughout shapes, also in the case of complex shapes (chair and motorcycle).

\paragraph{Few-shot learning.} 
Few-shot learning (FSL) aims to train a model that generalizes with limited
data. We conduct FSL ($N$-way $K$-shot learning) for the classification task on ModelNet40~\cite{sharma2016vconv} and ModelNet10~\cite{sharma2016vconv} benchmark datasets, where the model is evaluated on $N$ classes, and each class contains $N$ samples.   We use the same setting and train/test split as OcCo~\cite{wang2020unsupervised} and CrossPoint~\cite{afham2022crosspoint} and report the mean and standard deviation across 10 runs. Table~\ref{tab:few_shot} shows the FSL results on ModelNet40, where \ourmethod outperforms prior works in all the FSL settings in the DGCNN backbone. Our method with PointNet backbone performs slightly poorly in 5-way 10-short and 10-way 20-short settings compared to results of CrossPoint with PointNet.
\begin{table}[t]
	\centering
	\small
	\caption{Few-shot object classification results on ModelNet40.
We report mean and standard error over 10 runs. Top results of
each backbone is bold.}
	\label{tab:few_shot}
	\vspace{0.0cm}
	\setlength{\tabcolsep}{1.1mm}{
	\begin{tabular}{c l | c c c c}
		\toprule
		\multirow{2}{*}{Encoder} &  
		\multirow{2}{*}{Method}  & 
		\multicolumn{2}{c}{5-way} &  \multicolumn{2}{c}{10-way} \\
		\cline{3-6} 
 		\multicolumn{2}{c|}{} &  10-shot &  20-shot &  10-shot &  20-shot \\
		\toprule
		\multirow{6}{*}{PointNet} 
		& Rand                                      & 52.0 $\pm$ 3.8 & 57.8 $\pm$ 4.9 & 46.6 $\pm$ 4.3 & 35.2 $\pm$ 4.8 \\
		& Jigsaw~\cite{sauder2019self}              & 66.5 $\pm$ 2.5 & 69.2 $\pm$ 2.4 & 56.9 $\pm$ 2.5 & 66.5 $\pm$ 1.4 \\
		& cTree~\cite{sharma2020self} & 63.2 $\pm$ 3.4 & 68.9 $\pm$ 3.0 & 49.2 $\pm$ 1.9 & 50.1 $\pm$ 1.6 \\
        & OcCo~\cite{wang2020unsupervised} & 89.7 $\pm$ 1.9 & 92.4 $\pm$ 1.6 & 83.9 $\pm$ 1.8 & 89.7 $\pm$ 1.5 \\
        & CrossPoint~\cite{afham2022crosspoint}     & \bf 90.9 $\pm$ 4.8 & 93.5 $\pm$ 4.4 & 84.6 $\pm$ 4.7 & \bf 90.2 $\pm$ 2.2 \\
        & \ourmethod                                & 90.6 $\pm$ 4.0 & \bf 93.8 $\pm$ 3.2 & \bf 84.7 $\pm$ 3.6 & 90.1 $\pm$ 4.5 \\
		\midrule
		\multirow{6}{*}{DGCNN} 
		& Rand                                      & 31.6 $\pm$ 2.8 & 40.8 $\pm$ 4.6 & 19.9 $\pm$ 2.1 & 16.9 $\pm$ 1.5 \\
        & Jigsaw~\cite{sauder2019self}              & 34.3 $\pm$ 1.3 & 42.2 $\pm$ 3.5 & 26.0 $\pm$ 2.4 & 29.9 $\pm$ 2.6 \\
        & cTree~\cite{sharma2020self} & 68.4 $\pm$ 3.4 & 71.6 $\pm$ 2.9 & 42.4 $\pm$ 2.7 & 43.0 $\pm$ 3.0 \\
        & OcCo~\cite{wang2020unsupervised}          & 90.6 $\pm$ 2.8 & 92.5 $\pm$ 1.9 & 82.9 $\pm$ 1.3 & 86.5 $\pm$ 2.2 \\
        & CrossPoint~\cite{afham2022crosspoint}     & 92.5 $\pm$ 3.0 & 94.9 $\pm$ 2.1 & 83.6 $\pm$ 5.3 & 87.9 $\pm$ 4.2 \\
        & \ourmethod                                & \bf 93.6 $\pm$ 3.3 & \bf97.3 $\pm$ 2.0 & \bf 89.1 $\pm$ 1.4 & \bf 93.2 $\pm$ 3.4 \\
		\toprule
	\end{tabular}
	}
\end{table}

% %%%%%%%%%%%%%%%%%%%%%%%%%%%%%%%%%%%%%%%%%%%%%%%%%%%%%%%%%%%

\paragraph{Batch size.}
Contrastive methods mine negative examples from the mini-batch can suffer from performance drops when their batch size is not sufficiently large \cite{grill2020bootstrap}.
Because \ourmethod does not rely on negative examples, we expect it to be more robust to smaller batch sizes when compared to the contrastive approaches.
We empirically show the effect of different batch sizes by comparing the performance of \ourmethod with SimCLR \cite{chen2020simple}. 
We variate batch sizes from 8 to 48 during pre-training. 
Table~\ref{tab:batch} shows that SimCLR experiences degradation of performance when the batch size is 8, likely due to the small number of available negative samples.
By contrast, \ourmethod maintains a fairly stable performance throughout different batch size configurations.

%++++++++++++++++++++++++++++++++++++++
%++++++++++++++++++++++++++++++++++++++
\begin{table}[t]
	\centering
	\caption{Ablation study results of SoftClu by using DGCNN on ModelNet10 with different batch sizes during pre-training.}
	\label{tab:batch}
% 	\vspace{-.3cm}
% 	\resizebox{\columnwidth}{!}
	{%
	\begin{tabular}{l l c c c c c c}
		\toprule
		Encoder & Method & 8 & 16 & 24 & 32 & 40 & 48 \\
		\toprule
		 \multirow{2}{*}{PointNet} & SimCLR & 87.5 & 88.0 & 88.2 & 88.1 & 88.5 & 88.4  \\
		 & \ourmethod & 89.9 & 89.8 & 90.2 & 90.3 & 90.1 & 89.9 \\
		 \midrule
		 \multirow{2}{*}{DGCNN} & SimCLR & 88.6 & 89.3 & 89.4 & 89.7 & 89.7 & 90.1 \\
		 & \ourmethod  & 91.6 & 91.8 & 91.7 & 91.9 & 91.9 & 91.8 \\
		\bottomrule
	\end{tabular}
	}
\end{table}
%++++++++++++++++++++++++++++++++++++++
%++++++++++++++++++++++++++++++++++++++
%%%%%%%%%%%%%%%%%%%%%%%%%%%%%%%%%%%%%%%%%%%%%%%%%%%%%%%%%%%
\paragraph{Computation of soft-labels.}
We assess our strategy for soft-label assignment based on optimal transport (OT) by comparing it with a typical L2 distance-based approach on ModelNet40 and ModelNet10. 
Therefore, we assess \ourmethod by using $\Gamma$ computed with Eq.~(\ref{eq:gamma}) and by using the L2 approach in \cite{caron2018deep}.
Table~\ref{tab:ablation_gamma} shows that OT achieves the best performance on all the datasets with both PointNet and DGCNN encoders. 
This is due to the equal partition constraint in Alg.~\ref{alg:sinkhorn} which prevents solutions from being assigned to the same cluster and affecting the performance.

%++++++++++++++++++++++++++++++++++++++
%++++++++++++++++++++++++++++++++++++++
\begin{table}[!hbt]
	\centering
	\caption{Ablation study of \ourmethod on ModelNet40 and ModelNet10 with soft-labels computed with our approach (OT) and with a typical distance-based assignment (L2).}
	\label{tab:ablation_gamma}
% 	\vspace{-.3cm}
	\begin{tabular}{l l c c}
		\toprule
		\multirow{2}{*}{Dataset} & 
		\multirow{2}{*}{Encoder} &
		\multicolumn{2}{c}{Accuracy} \\
		\cline{3-4} 
 		\multicolumn{2}{c}{} & L2 & OT \\
		\toprule
		\multirow{2}{*}{ModelNet10} 
		& PointNet & 91.5 & \bf93.4 \\
		& DGCNN & 94.1 & \bf94.8 \\
		\midrule
		\multirow{2}{*}{ModelNet40} 
		& PointNet & 86.5 & \bf90.3 \\
		& DGCNN & 90.4 & \bf91.9 \\
		\toprule
	\end{tabular}
	
\end{table}
%++++++++++++++++++++++++++++++++++++++

\paragraph{\ourmethod with Transformer backbone.}
Following~\cite{liu2022masked} setups, we also provide the results with the recent Transformer backbone provided by \cite{liu2022masked}. As shown in Tab.~\ref{tab:nbk}, \ourmethod also achieves competitive result.
%++++++++++++++++++++++++++++++++++++++
\begin{table}[!hbt]
	\centering
 \vspace{-.13cm}
	\caption{Classification results with a Transformer backbone on ModelNet40.}
	\label{tab:nbk}
% 	\resizebox{\columnwidth}{!}
	{%
\begin{tabular}{l l c c c c c c}
	\toprule
	Encoder &  \ourmethod & PointViT-OcCo~\cite{wang2020unsupervised} & Point-BERT~\cite{yu2022point} & MaskPoint~\cite{liu2022masked} \\
	\hline
	\ourmethod & \bf 93.8     & 92.1      & 93.2   & \bf 93.8 \\
 \bottomrule
	\end{tabular}
	}
\end{table}
%++++++++++++++++++++++++++++++++++++++

\end{document}